\documentclass[letterpaper,10 pt,journal,twoside]{IEEEtran}

\usepackage{amsthm}
\newtheorem{remark}{Remark}

\usepackage[dvipsnames]{xcolor}

\usepackage{graphics}
\usepackage{amsmath} 
\usepackage{amssymb}  
\usepackage{tablefootnote}
\usepackage{threeparttable}
\usepackage{tikz}

\usepackage{amsmath}
\usepackage{amssymb}

\usepackage{mathrsfs}
\usepackage{hyperref}       
\usepackage{url}            
\usepackage{booktabs}       
\usepackage{amsfonts}       
\usepackage{nicefrac}       
\usepackage{lipsum}
\usepackage{tabularx} 
\usepackage{amsmath}  
\usepackage{graphicx,dblfloatfix} 
\usepackage{multirow}

\usepackage{cite} 
\usepackage{titlesec}
\usepackage{mathtools, amssymb, nccmath}
\usepackage{algorithm}
\usepackage{algorithmic}
\usepackage{multicol}

\usepackage{enumitem}
\usepackage{float}
\usepackage{mathtools}
\usepackage{amsmath,amsfonts,amssymb}
\usepackage{cmll}
\usepackage{tensor}
\usepackage{booktabs}
\usepackage{tabularx}
\usepackage{makecell}

\usepackage[font=small,skip=5pt]{caption}
\usepackage[]{xcolor}

\newcolumntype{Z}{ >{\centering\arraybackslash}X }

\newcommand{\crff}{\,\overline{\!\times\!}{}^{\,*}}

\newcommand{\M}{\mathcal{M}}
\newcommand{\F}{\mathbf{F}}
\newcommand{\I}{\mathcal{I}}

\newcommand{\crf}{\times^*}
\newcommand{\crm}{\times}
\newcommand{\T}{^\top}

\newcommand{\R}{\mathbb{R}}
\newcommand{\C}{\vC}

\renewcommand{\H}{\vH}

\newcommand{\XM}[2]{ \tensor[^{#1}]{\mathbf{X}}{_{#2}}}
\newcommand{\XMT}[2]{ {}^{#1}\mathbf{X}\T_{#2}}

\renewcommand{\L}{\mathscr{L}}

\newcommand{\greekvec}[1]{\boldsymbol{#1}}

\newcommand{\B}{\mathbf{B}{}}
\renewcommand{\C}{\mathbf{C}}
\renewcommand{\M}{\mathbf{M}}
\newcommand{\X}{\mathbf{X}}
\newcommand{\J}{\mathbf{J}}

\newcommand{\bzero}{\mathbf{0}}

\renewcommand{\I}{\mathbf{M}}
\newcommand{\g}{\mathbf{g}}
\newcommand{\q}{{\mathsf{q}}}

\renewcommand{\u}{\mathbf{u}}

\newcommand{\f}{\mathbf{f}}

\renewcommand{\v}{\mathbf{v}}
\renewcommand{\t}{\mathbf{t}}

\renewcommand{\H}{\mathbf{H}}
\renewcommand{\a}{\mathbf{a}}

\newcommand{\btau}{\boldsymbol{\tau}}
\newcommand{\qd}{{\dot{\q}}}
\newcommand{\qdd}{{\ddot{\q}}}

\newcommand{\bq}{\mathbf{q}}
\newcommand{\bqd}{\dot{\bq}}
\newcommand{\bqdd}{\ddot{\bq}}

\newcommand{\Ppsi}{\greekvec{\Psi}}
\newcommand{\Ppsid}{\dot{\Ppsi}}
\newcommand{\Ppsidd}{\ddot{\Ppsi}}

\newcommand{\taubar}{\greekvec{\tau}}

\newcommand{\red}[1]{{\color{black}#1}}
\newcommand{\purple}[1]{{\color{purple}#1}}
\newcommand{\blue}[1]{{\color{blue}#1}}

\renewcommand{\S}{\mathbf{S}}

\newcommand{\Sd}{\,\dot{\!\S}}

\newcommand{\Sring}{\,\mathring{\!\S}}
\newcommand{\gSring}{\mathring{\mathsf{S}}}
\newcommand{\gSd}{\dot{\mathsf{S}}}

\newcommand{\p}[1]{{p(#1)}}

\DeclareMathOperator{\ID}{ID}

\newcommand{\vomega}{\boldsymbol{\omega}}
\newcommand{\vv}{\boldsymbol{v}}

\newcommand{\Jac}[2]{\frac{\partial #1}{\partial#2}}

\newcommand{\Uupd}{\dot{\boldsymbol{\Upsilon}}}

\newcommand{\pluseq}{\mathrel{+}=}

\newcommand{\IC}{\breve{\M}}
\newcommand{\MC}{\breve{\M}}
\newcommand{\fC}{\breve{\f}}

\newcommand{\BC}{\breve{\B}}

\newcommand{\BCT}[1]{\BC{}_#1\T}

\usepackage{setspace}

\newcommand{\glob}[1]{\mathsf{#1}}

\renewcommand{\u}{\mathbf{u}}

\newcommand{\gA}{\glob{A}}
\newcommand{\gB}{\mathsf{B}}
\newcommand{\gBC}{\breve{\gB}}

\newcommand{\gF}{\mathsf{F}}
\newcommand{\gf}{\mathsf{f}}

\newcommand{\gI}{\mathsf{M}}

\newcommand{\gIC}{ \breve{\mathsf{M}} }
\newcommand{\gMC}{ \breve{\mathsf{M}} }

\newcommand{\gM}{\mathsf{M}}

\newcommand{\gS}{\mathsf{S}}

\newcommand{\gV}{\mathsf{V}}

\newcommand{\gv}{\mathsf{v}_{\rm J}}

\newcommand{\eye}{\mathsf{1}}

\newcommand{\gY}{\mathsf{y}}

\newcommand{\gUpd}{\dot{\mathsf{\Upsilon}}}

\newcommand{\bone}{\mathbf{1}}

\newcommand{\vf}{\boldsymbol{f}}
\graphicspath{{./}{../}}

\usepackage{scalerel}

\begin{document}

\title{Adapting Rigid-Body Dynamics Derivatives\\ for Constraint Embedding Closed-Chain Models}

\author{Daniel J.~Volpi and Patrick M.~Wensing
\thanks{Manuscript received: February 1st, 2026; Revised May 8th, 2026; Accepted June 8th, 2026.}
\thanks{This paper was recommended for publication by Editor L.~Pallottino upon evaluation of the Associate Editor and Reviewers’ comments.
This work was supported by the National Science Foundation (NSF) grant CMMI-2220924 through a subaward to the University of Notre Dame and via the NSF Graduate Research Fellowship Program under grant DGE-2236418.} 
\thanks{Daniel J.~Volpi and Patrick M.~Wensing are with the Dept. of Aerospace \& Mechanical Engineering, University of Notre Dame, IN-46556, USA. \href{mailto:dvolpi@nd.edu}{dvolpi@nd.edu}, \href{mailto:pwensing@nd.edu}{pwensing@nd.edu}.}
\thanks{Digital Object Identifier (DOI): see top of this page.}\\[-5ex]~
}

\markboth{IEEE Robotics and Automation Letters. Preprint Version. Accepted June, 2026. DOI: 10.1109/LRA.2026.3709575}
{Volpi and Wensing: Rigid-Body Dynamics Derivatives for Constraint Embedding Closed-Chain Models} 
\newcommand{\Q}{\mathcal{Q}}

\maketitle 

\begin{abstract}
This paper extends an existing algorithm for the first-order derivatives of rigid-body dynamics to the case of closed-chain kinematic systems modeled using constraint embedding. Many standard dynamics algorithms apply to both open-chain and constraint-embedded models, but existing efficient derivative methods assume joint velocity effects are locally configuration invariant.
We remove this assumption and derive adapted algorithms that extend dynamics derivatives to more general joint types, including those arising in constraint-embedded closed-chain models. Our results compare conventional pin-joint robot models with more complete actuation models that capture local closed chains. We show that the additional terms introduced by these generalizations have low computational impact when modeling actuation kinematics alone, but can incur higher cost when additional rigid bodies, such as motor rotors, are included in the actuation chain, or when considering non-local loops. Overall, these results enable more accurate dynamics computations for constraint-embedded actuation submechanisms to be adopted in model-predictive control and differentiable simulation.
\end{abstract}


\section{Introduction}

\IEEEPARstart{T}{he} accurate and efficient computation of rigid-body dynamics derivatives plays
a key role in modern optimization-based control strategies for robotic systems.
Model predictive control (MPC) frameworks~\cite{wensing2023optimization, grandia2023perceptive, jordana2025structure, mastalli2023inverse, li2024cafe}
require dynamics derivatives for rapid trajectory optimization. For learning-based
control and system identification tasks, gradients of the dynamics are used in
differentiable physics simulations~\cite{toussaint2018differentiable,sutanto2020encoding}.
However, the most efficient rigid-body dynamics derivatives
algorithms have been specialized to open-chain systems with a restricted set of joint types. 
With the increasing
prevalence of robots with complex actuation systems employing local closed chains~\cite{sim2022tello, chignoli2021humanoid, liu2022design},
there is a growing need for the accurate and efficient computation of rigid-body dynamics derivatives for closed-chain
kinematic structures.

Classically, Lagrange multiplier methods~\cite{Featherstone08, carpentier2021proximal}
are used to treat the forward dynamics of closed-chain mechanisms, departing from efficient recursive structures that are enjoyed by open-chain models. By contrast, constraint embedding~\cite{jain2012multibody} is a powerful modeling structure that enables many open-chain algorithms (e.g., both forward and inverse dynamics) to be applied to closed-chain systems. It accomplishes this by taking localized kinematic loops and clustering the involved bodies into aggregate bodies. Following this clustering process, the connectivity of the aggregate bodies can be made into a tree, where existing recursive approaches still apply in a generalized sense. While originally developed following the lineage of Jain's spatial operator algebra \cite{jain2012multibody}, recent work has provided an alternative perspective more closely connecting the approach to Featherstone-style \cite{Featherstone08} propagation methods \cite{chignoli2025propagation}. 
Constraint embedding has seen a resurgence in recent years, in particular, for
serial/closed-chain hybrid robots~\cite{kumar2020analytical, kumar2022modular}.
Beyond accuracy considerations for simplified actuation models, neglecting closed-chain structures in actuation can limit performance~\cite{boukheddimi2023investigations}. While many open-chain algorithms extend readily to constraint-embedding models, derivative algorithms rely on assumptions that do not hold in this setting.

Many approaches have been considered for computing derivatives of rigid-body dynamics algorithms. Numerical methods of finite difference and
complex-step differentiation~\cite{cossette2020complex} come with round-off or increased overhead, respectively, while automatic
differentiation~\cite{giftthaler2017automatic} automates the application of the chain rule in forward-mode or reverse-mode fashions. These approaches tie the computation of derivatives of a function to its original computation graph, which, while general, is not always the most efficient. By contrast, analytical methods may employ identities or explore other computation graphs, enabling efficiency beyond general-purpose approaches. Extensions to original formulations \cite{JainRodriguez93} have been refined and implemented in libraries such as Pinocchio~\cite{carpentier2019pinocchio, carpentier2018analytical}. Closed-form expressions for robots with general multi-DoF Lie group joints were developed in \cite{Singh22}, enabling further efficiency gains.
Alternative analytical formulations
have also been proposed~\cite{Bobrow01, ayusawa2018comprehensive}. 

Despite these advances, existing efficient recursive algorithms for rigid-body dynamics derivatives \cite{JainRodriguez93, carpentier2018analytical, Singh22} have focused on open-chain joint models where the joint motion subspace matrices (denoted as $\S_i$ in past work and in Featherstone’s text \cite{Featherstone08}) are constant in body coordinates. 
For single-degree of freedom joints, this is the same as assuming the joint twists are constant in body coordinates.
While the chain-rule derivation in \cite{carpentier2018analytical} permits the appearance of additional terms associated with configuration dependence in $\S_i$, more efficient algorithmic structures that depart from explicit chain-rule propagation \cite{JainRodriguez93, carpentier2019pinocchio, Singh22} have been specialized to this configuration-invariant setting. This specialization is appropriate for standard joints (e.g., revolute, prismatic, floating base), but it fails for more general joint types, including those arising in constraint-embedded closed-chain models. 

As our primary contribution, we identify and remove this
configuration-invariance assumption present in existing efficient derivative algorithms,
deriving adapted formulas that apply to the more general joint types needed for constraint-embedding models.
We then implement the algorithm as an open-source contribution \cite{GRBA} to the Generalized Rigid-Body
Dynamics Algorithms (GRBDA) library \cite{chignoli2025propagation}. We demonstrate correctness through comparison with
complex-step and provide comprehensive performance
analysis comparing conventional pin-joint models with complete actuation models
that capture local closed chains. Our results show that modeling actuation
kinematics alone introduces low computational overhead, while including
additional rigid bodies such as motor rotors incurs higher computational
costs, which must be weighed in practical applications against the increased accuracy of the model. We focus on the derivatives of inverse dynamics, deferring to existing results \cite{JainRodriguez93, carpentier2018analytical} that relate the derivatives of forward dynamics to those of inverse dynamics.

\section{Background}
\label{sec:background}

This section reviews spatial vector notation \cite{Featherstone08} for computing the inverse dynamics of open-chain mechanisms. A central motivation of the manuscript is to build out an algorithm for dynamics derivatives of constraint-embedding closed-chain models \cite{jain2012multibody}. Here, we focus primarily on open-chain systems for clarity of prose, since once we discharge the configuration invariance assumption discussed, the architecture of the derivative computations for constraint embedding models will take an identical form. 

\subsection{Rigid-Body Dynamics}
We consider an open-chain rigid-body system with $N_B$ bodies, $N$ joints, and $n$ total degrees of freedom (DoFs). We partition the configuration manifold $\Q$ as $\Q = \Q_{1} \times ... \times \Q_{N}$, where each $\Q_{i}$ represents the configuration manifold of joint $i$. The number of DoFs of joint $i$ is denoted by $n_{i} $, and $n=\sum_{i=1}^{N} n_i$. The state variables associated with each joint are a representation of the configuration $\bq_{i}$ and the generalized velocity vector $\bqd_i\in\mathbb{R}^{n_{i}}$, while the control variable is the generalized force/torque vector $\taubar_{i} \in \mathbb{R}^{n_{i}}$. We stack these quantities across all joints into $\q$, $\qd$, and $\btau$.

With this convention, each $\bqd_i$ will uniquely specify the time rate of change for the configuration $\bq_i \in \Q_i$ without strictly being its time derivative. Specifically, the rate of change in the configuration representation will be given by
\begin{equation}
\frac{\rm d}{ {\rm d} t} \bq_i = \sum_{j=1}^{n_i} X_{ij}(\bq_i) \dot{q}_{ij} 
\end{equation}
where each function $X_{ij}$ details the configuration change created from joint $i$, velocity component $j$. 
For example, for a spherical joint, $\bq_i$ could be selected as a quaternion for the joint, while $\bqd_i$ could be selected as the relative angular velocity between the bodies. In this case, $X_{ij}$ would represent the quaternion rates from rotations about the body $x$, $y$, and $z$ axes. With these conventions, for a vector-valued function $\vf : \Q_i \rightarrow \R^m$, when we write $\frac{\partial \vf}{\partial \bq_i} $ we will consider it to mean
\begin{equation}
\frac{\partial \vf}{\partial \bq_i} := \begin{bmatrix} \L_{X_{i,1}} \vf & \cdots & \L_{X_{i,n_i}} \vf \end{bmatrix}
\label{eq:lie_derivs}
\end{equation}
as a matrix of Lie derivatives $\L_{X_{i,j}} \vf$ \cite[Ch.~7, Sec.~2.1]{murray1994mathematical}. The benefit of this approach is that it decouples the derivatives from the choice of configuration representation (e.g., the above would give the same result for the spherical joint if the joint configuration was represented with a rotation matrix vs.~a quaternion). For complete details on this Lie-theoretic approach to robot derivatives, see \cite{Singh22, singh2024second}.

The Inverse Dynamics (ID) is given by 
 \begin{align}
    \taubar &= \H(\q)\qdd+ \C(\q,\qd)\qd + \g(\q)
    \label{inv_dyn} \\
&= \textrm{ID}({\rm model},\q,  \qd, \qdd)
    \label{inv_dyn_model}
\end{align}
where $\H \in \R^{n \times n}$ is the mass matrix, $\C \in \R^{n \times n}$ is the Coriolis matrix, and  $\g \in \R^{n}$ is the vector of generalized gravitational forces.
For a fixed configuration and velocity, inverse dynamics computes $\taubar$ given $\qdd$, while forward dynamics computes $\qdd$ for a given $\taubar$:
\begin{align}
    \qdd &= \H^{-1}(\q)\left( \taubar -  \C(\q,\qd)\qd - \g(\q) \right)
    \label{fwd_dyn2} \\
    &= \textrm{FD}({\rm model},\q, \qd, \taubar)
    \label{for_dyn_model}
\end{align}
The first-order partial derivatives of forward dynamics can be written by exploiting the derivatives of ID as~\cite{JainRodriguez93,carpentier2018analytical}:
  \begin{equation}
      \frac{\partial~\textrm{FD}}{\partial \boldsymbol u}\biggr\rvert_{\q_{0},  \qd_{0}, \taubar_{0}} = -\H^{-1}(\q_{0}) \frac{\partial~  \textrm{ID} }{\partial \boldsymbol u}\biggr\rvert_{\q_{0},  \qd_{0},  \qdd_{0}}
      \label{jain_FO_eqn}
  \end{equation}
where $\u \in \{\q,\qd\}$. Owing to this relationship, we hereafter focus on the derivatives of inverse dynamics alone.


\subsection{Spatial Vectors}
\label{iden_defn}

{\bf Notation:} Spatial vectors are 6D vectors that combine the linear and angular aspects of a rigid-body motion or net force~\cite{Featherstone08}. Spatial vectors are denoted with  lower-case bold letters (e.g., $\a$), while matrices are denoted with capitalized bold letters (e.g., $\boldsymbol{A}$). 
Spatial vectors are usually expressed in either the ground coordinate frame or a body coordinate (local coordinate) frame. For example, the spatial velocity ${}^{k}\v_{k} \in \R^{6}$ of a body $k$ expressed in the body frame is
${}^{k}\v_{k} = \begin{bmatrix}
           {}^{k}\vomega_{k}\T &
           {}^{k}\vv_{k}\T
         \end{bmatrix}\T$
where ${}^{k}\vomega_{k}$ $\in \mathbb{R}^{3}$ is the angular velocity expressed in a coordinate frame fixed to the body, while ${}^{k}\vv_{k}$ $\in \mathbb{R}^{3}$ is the linear velocity of the origin of the body frame. 
When the frame used to express a spatial vector is omitted, the body frame is assumed.

 A spatial cross product between two motion vectors ($\v$,$\u$), written as $(\v \times) \u$, is given by \eqref{cross_defn}. This operation gives the time rate of change of $\u$, when $\u$ is moving with a spatial velocity $\v$.  
For a Cartesian vector, the matrix $\vomega \times$ is the 3D cross-product operator. A spatial cross product between a motion and a force vector is written as $(\v \times^*) \f$, as defined in \eqref{cross_defn}.
\begin{align}
    \v \times = \begin{bmatrix}
       \vomega  \times & \bf{0}  \\
       \vv \times &  \vomega \times
    \end{bmatrix}
    ~~~ & ~~~
    \v \times^* = \begin{bmatrix}
        \vomega \times & \vv \times  \\
        \bf{0} &  \vomega \times
    \end{bmatrix}
    \label{cross_defn}
\end{align}

\noindent An operator $\crff$ is  defined by swapping the order of the cross product, such that $(\f \crff) \v = (\v \times^*) \f$.  

\begin{figure}[tb]
\centering
\includegraphics[width=.9 \columnwidth]{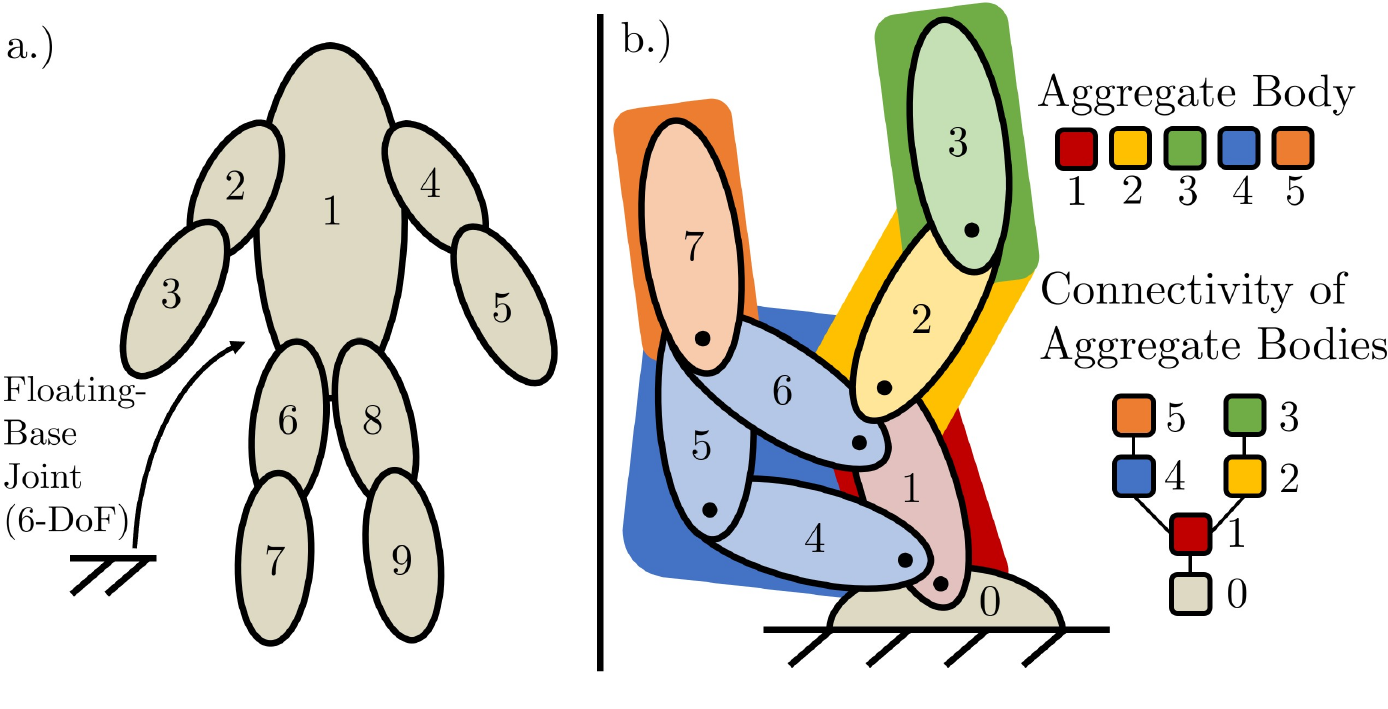}
\caption{Body numbering and notation examples for (a) an open-chain system and (b) a constraint-embedding model applied to a closed-chain system.}
\label{fig:connectivity}
\vspace{-5px}
\end{figure}

{\bf Connectivity:} An open-chain kinematic tree with serial or branched connectivity (Fig.~\ref{fig:connectivity}(a)) is considered with $N_B=N$ links connected by joints, each with up to 6 DoF. Body $i$'s parent toward the root of the tree is denoted as $\p{i}$. We define $i \preceq j$ if joint $i$ is in the path from body $j$ to the base.



The spatial velocities of the neighboring bodies in the tree are then related by $\v_i = \XM{i}{\p{i}}(\bq_i)\,  \v_\p{i} + \S_i(\bq_i) \bqd_i$, where $\S_i(\bq_i)$ is the joint motion subspace matrix for joint $i$ \cite{Featherstone08} and $\XM{i}{\p{i}}(\bq_i)$ denotes the spatial transform between frames. Past derivatives derivations (e.g., \cite{JainRodriguez93, Singh22}) assumed $\S_i(\bq_i)$ to be fixed in local coordinates such that its dependence on $\bq_i$ could be dropped. 
The velocity $\v_i$ can also be written as the sum of joint velocities over predecessors: $\v_i = \sum_{l \preceq i} \XM{i}{l} \S_{l} \bqd_{l}$.

Spatial accelerations of neighboring bodies are related by
\begin{equation}
\a_i = \XM{i}{\p{i}} \a_{\p{i}} + \Sd_i \bqd_i + \S_i \bqdd_i
\end{equation}
where we follow the derivative notation of Featherstone \cite{Featherstone08} such that
\begin{equation}
\Sd{}_i = \v_i \times \S_i + \Sring_i
\end{equation}
gives the absolute derivative of $\S_i$ (derivative taken inertially, but expressed locally), and $\mathring\S_i = \frac{\rm d}{{\rm d}t} \S_i$ is the derivative taken in the local coordinates. Past derivations \cite{JainRodriguez93, Singh22} and the implementation of \cite{carpentier2018analytical} assumed that this term ($\Sring_i$) was zero.


{\bf Dynamics:} The spatial equation of motion~\cite{Featherstone08} is given for body $k$ as
$\f_{k}= \I_{k}\a_{k} + \v_{k} \times^*\I_{k} \v_{k}$, 
where $\f_{k}$ is the net spatial force on body $k$ and $\I_{k}$ is its spatial inertia~\cite{Featherstone08}. Instead of treating gravity as an external force, a common trick is to accelerate the base upwards opposite of the gravitational acceleration ($\a_{0} = -\a_{g}$), providing the acceleration of body $k$ as
\vspace{-1ex}
\begin{equation}
    \a_{k} = \textstyle\sum_{l \preceq k} \XM{k}{l} \big( \S_{l} \bqdd_{l} + \Sd_{l} \bqd_{l} \big)+\XM{k}{0} \a_{0}\,
    \label{spatial_acc}
\end{equation}
Via the RNEA~\cite{Featherstone08}, $\btau_i = \S_i^{T}\fC_i $, where $\fC_i=\sum_{k \succeq i} \f_{k}$ is the spatial force transmitted 
across joint $i$. Hereafter, we ignore gravity, since its influence can be addressed in all final algorithms by adding this acceleration bias. 

{\bf Constraint Embedding Models:}  
The main idea behind constraint embedding \cite{jain2012multibody} is that bodies involved in each local kinematic loop can be grouped together into an aggregate body (i.e., a cluster of bodies) such that the connectivity of the aggregate bodies is that of a tree (Fig.~\ref{fig:connectivity}(b)). In these settings, the velocity propagation $\v_i = \XM{i}{\p{i}}(\bq_i)\,  \v_\p{i} + \S_i(\bq_i) \bqd_i$ is generalized so that $\v_i$ represents the concatenation of all the velocities of the rigid bodies in the $i$-th aggregate body, and $\S_i$ encodes how all of the rigid bodies in the $i$-th aggregate body move relative to the bodies in its parent. (See \cite{jain2012multibody} for the original development, and \cite{chignoli2025propagation} for an alternate perspective). 

While $\S_i$ being configuration invariant for open-chain models is the norm (e.g., for revolute, prismatic, and floating-base joints), it is uncommon for the generalized joint models appearing in constraint embedding. This is due to the configuration-dependency of many input-output relationships for common mechanisms (e.g., non-parallelogram four-bar linkages) making $\S_i(\bq_i)$ highly mechanism specific. 

As a concrete example, consider aggregate body $4$ in Fig.~\ref{fig:connectivity}.
We consider a spanning tree associated with breaking the constraint between bodies $6$ and $1$, making $\bq_4$, $\bq_5$, $\bq_6$ refer to the spanning tree, while we use $\overline{\bq}_4$ to refer to the configuration for the aggregate body. We could choose $\overline{\bq}_4$, for example, by picking one specific joint angle in the chain. We choose $\overline{\bq}_4 = \bq_4$ here for illustrative purposes. A loop constraint on the spanning tree can be expressed as
\begin{equation}
\bf{0} = \begin{bmatrix} \mathbf{J}_4 & \mathbf{J}_{5} & \mathbf{J}_6 \end{bmatrix} \left[ \begin{smallmatrix} \bqd_4 \\ \bqd_5 \\ \bqd_6\end{smallmatrix} \right]
\label{eq:constriant_jacobian}
\end{equation}
which then gives the spanning tree joint velocities as
\begin{equation}
\left[ \begin{smallmatrix} \bqd_4 \\ \bqd_5 \\ \bqd_6 \end{smallmatrix}\right] = \begin{bmatrix} 1 \\ -\begin{bmatrix} \J_5 & \J_6 \end{bmatrix}^{-1} \J_4 \end{bmatrix} \bqd_4 = \mathbf{G}(\overline{\bq}_4) \dot{\overline{\bq}}_4
\label{eq:G}
\end{equation}
with $\mathbf{G}(\overline{\bq}_4)$ defined based on the equivalent quantity in the middle expression. This then gives the relative velocities of all bodies in aggregate body $4$ relative to the parent as
\begin{equation}
\begin{bmatrix} \v_4 - \v_1\\ \v_5 -\v_1\\ \v_6-\v_1 \end{bmatrix} = \underbrace{ \begin{bmatrix} \S_4 & \bzero & \bzero \\ \XM{5}{4} \S_4 & \mathbf{0} & \bzero \\ \XM{6}{4} \S_4 & \XM{6}{5} \S_5 & \S_6 \end{bmatrix} \mathbf{G} (\overline{\bq}_4) }_{\triangleq \overline{\S}_4(\overline{\bq}_4)} \dot{\overline{\bq}}_4  
\label{eq:S}
\end{equation}
which defines the equivalent of the conventional joint-motion subspace matrix for a more general joint model used in constraint embedding.

It is not scalable to carry out this process for more complex loops by hand. To address this challenge, the open-source GRBDA library from \cite{chignoli2025propagation} enables automatically forming $\S_i(\bq_i)$ from a spanning tree model of an aggregate body. It does this by following the same steps as the example above, namely by a) having the user note independent coordinates, b) taking in a constraint function to construct a constraint Jacobian \eqref{eq:constriant_jacobian}, and c) following the same construction as \eqref{eq:G} and \eqref{eq:S} to construct $\S_i(\bq_i)$. The URDF+ parser \cite{Chignoli24_urdfplus} further automates this process by automatically grouping rigid bodies into aggregate bodies and constructing the constraint function for planar loops. Overall, this process results in nonlinear terms in $\S_i(\bq_i)$ due to $\sin$/$\cos$ terms in the spatial transforms and matrix inverses of blocks of the constraint Jacobian.
The remainder of the development applies equally to open-chain and~constraint embedding models, and we make no further distinction. 

\newcommand{\be}{\begin{equation}}
\newcommand{\ee}{\end{equation}}
\renewcommand{\crm}[1]{(#1 \times)}
\newcommand{\stimes}{\times^*}
\renewcommand{\crf}[1]{(#1 \stimes)}
\renewcommand{\otimes}{\,\overline{\!\times\!}\,^*}
\newcommand{\icrf}[1]{(#1 \otimes)}
\newcommand{\BlkDiag}{{\rm BlkDiag}}
\newcommand{\Sum}{\mathsf{\Sigma}}

\newcommand{\gradq}{\nabla_{\q}}
\newcommand{\grad}{\nabla}

\newcommand{\Sumt}{\overline{\Sum}}
\newcommand{\SumtT}{\Sumt{}\T}

\renewcommand{\red}[1]{{\color{red}#1}}
\renewcommand{\purple}[1]{{\color{purple}#1}}
\newcommand{\Om}{\Omega}

\renewcommand{\gS}{\mathsf{S}}

\newcommand{\gPsi}{\mathsf{\Psi}}
\newcommand{\gPsid}{\dot{\gPsi}}
\newcommand{\gPsidd}{\ddot{\gPsi}}

\newcommand*\circled[1]{\tikz[baseline=(char.base)]{
            \node[shape=circle,draw,inner sep=2pt] (char) {\small #1};}}

\newcommand{\colored}[2]{{\color{#1}#2}}
\newcommand{\colornumber}[3]{ \colored{#2}{ #3}}
\renewcommand{\blue}[1]{{\color{blue}#1}}

\newcommand{\bv}{\mathbf{v}}

\section{Global Factorization of the RNEA}
\label{sec:global}

To simplify our analysis, we work with a global factorization of the inverse dynamics equations, as in \cite{JainRodriguez93}. This greatly simplifies the derivation compared to the approach of \cite{Singh22}, which proceeded body-by-body. Instead, by working with equations for the system as a whole, we will mirror the more compact development of \cite{JainRodriguez93} while extending its applicability. We denote body/joint-level quantities with bolds (e.g., $\bq_i$, $\S_i$), and employ a non-bold font to refer to system-level quantities that stack all bodies/joints (e.g., $\q$, $\gS$).

We collect the joint free-mode matrices for all joints as
\be
\gS \triangleq \textrm{BlkDiag}(\S_1, \ldots, \S_N) \in \mathbb{R}^{6 N_B \times n}
\ee

Likewise, we can stack the joint velocities of all joints
\be
\gv \triangleq \gS \, \qd =\begin{bmatrix} \S_1 \bqd_1 \\ \vdots \\ \S_N \bqd_N\end{bmatrix} \in \mathbb{R}^{6 N_B}
\ee
and the velocities of all links (each expressed in body frames)
\be
\gV \triangleq \Sum \,  \gS \, \qd = \left[\begin{smallmatrix} \bv_1 \\ \vdots \\ \bv_{N_B} \end{smallmatrix}\right] \in \mathbb{R}^{6 N_B}
\ee
where the sum operator $\Sum \in \mathbb{R}^{6 N_B \times 6 N_B} $ is given by a block matrix that sums over all predecessors such that
\be
\gV = \Sum \gv
\ee

For example, in a tree with predecessor array $p = [0,1,2, 1]$ the matrix would take the form
\be
\scaleobj{.89}{
\Sum = \begin{bmatrix} {}^1 \X_1 & 0 & 0 & 0 \\ {}^2 \X_1 & {}^2 \X_2 & 0 &0 \\ {}^3 \X_1 & {}^3 \X_2 & {}^3 \X_3 & 0 \\ {}^4 \X_1 & 0 & 0 & {}^4 \X_4  \end{bmatrix} = \begin{bmatrix} \bone & 0 & 0 & 0 \\ {}^2 \X_1 & \bone & 0 &0 \\ {}^3 \X_1 & {}^3 \X_2 & \bone & 0 \\ {}^4 \X_1 & 0 & 0 & \bone  
\end{bmatrix}
}
\ee
Owing to this structure, we define $\Sumt = \Sum- \eye$.

We consider the cross product matrix of a global quantity by stacking spatial cross product matrices on the diagonal
\be
(\gV \times) = {\rm BlkDiag} \left( (\bv_1 \times) , \ldots,  (\bv_N\times \right) ) \in \mathbb{R}^{6 N_B \times 6 N_B}
\ee
with $\gV \times^* $ and $\gF \crff$ defined similarly. Since ${}^i \dot{\X}_j = ({}^i \bv_j - {}^i \bv_i) \times {}^i \X_j = - \bv_i \times {}^i \X_j + {}^i \X_j \times \bv_j$ it follows that
\be
\dot{\Sum} = -(\gV\times)\Sum + \Sum (\gV \times)
\label{eq:SumDot}
\ee
This relationship will be key for later developments.


We consider the spatial inertia for all the links (each expressed in their body frames) as:
\be
\gI = {\rm BlkDiag} \left( \I_1 , \ldots,  \I_N \right) \in \mathbb{R}^{6 N_B \times 6 N_B}
\ee
With these definitions, we can write down accelerations as
\begin{align}
\gA &\triangleq \Sum \, \left( \gS \, \qdd + \crm{\gV}\, \gv + \gSring \qd \right) \\
  &= \Sum \,(\gS \, \qdd+\gSring \, \qd) - \Sum \, \crm{\gv} \, \gV
\end{align}

The net force on each link is given in stacked form by
\be
\gf \triangleq \gI  \gA + \crf{\gV} \gI  \gV
\ee
with the forces at the joints (i.e., the sum of the net forces over each subtree) given by
\be
\gF \triangleq \Sum\T \gf = \Sum\T \left[ \gI\, \gA \,+ \, \crf{\gV} \,\gI \,\gV \,\right]
\ee
Finally, the torque along the free modes of the joints is
\begin{align}
\scaleobj{.85}{\btau} &\triangleq \scaleobj{.85}{\gS\T \gF} \\
 &= \scaleobj{.85}{\gS\T \Sum\T \left[ \gI \, \Sum \, \gS \, \qdd \, + \gI \, \Sum \, \gSring \, \qd \, + \,\crf{\gV}\, \gI \, \gV - \gI \, \Sum \, \crm{\gv} \, \gV \, \right]} \label{eq:ID_eq}
\end{align}
\section{Contribution: First-Order Derivatives Applicable to Constraint-Embedding Models}
\label{sec:derivatives}

\subsection{Building Blocks}

Consider a purely configuration-dependent function $\vf(\q) : \Q \rightarrow \R^m$.
For any given joint, the matrix of Lie derivatives associated with that joint \eqref{eq:lie_derivs} is alternatively defined by
\be
\frac{\partial \vf }{\partial \bq_i} = \frac{\partial}{\partial \dot{\bq}_i}\left[ \frac{{\rm d}}{{\rm d} t} \vf(\q) \right]
\ee
Likewise, for the full configuration $\q$, we associate a matrix of Lie derivatives by stacking these up
\be
\frac{\partial \vf }{\partial \q} := \nabla_\q \vf  = \left [ \frac{\partial \vf }{\partial \bq_1} \cdots \frac{\partial \vf}{\partial \bq_N} \right] 
\ee

\noindent Consider now a fixed vector $\gY \in \mathbb{R}^{6 N_B}$ and the fixed block diagonal inertia matrix $\gI$. We have the following due to \cite{JainRodriguez93}:
\begin{align}
\Sum (\gY \times) \Sum &=  ( \Sum \gY \times) \Sum - \Sum (\Sumt \gY \times)  \label{eq:I1}\\[.4ex]
\Sum\T \icrf{\gY} \Sum &= \icrf{ \Sum\T \gY } \Sum + \Sumt{}\T \icrf{\Sum\T \gY}  \label{eq:I2}\\[.4ex]
\Sum\T \gI \Sum &= \Sum\T \gIC + \gIC \Sum - \gIC  \label{eq:I3}
\end{align}
where $\gIC \in \mathbb{R}^{6 N_B \times 6 N_B}$ is the block diagonal matrix of composite inertias with diagonal blocks $\MC_i = \sum_{k\succeq i} {}^k \X_i\T \M_k {}^k \X_i $.

One of the main sources of configuration dependence in \eqref{eq:ID_eq} is through the transformation and summation operator $\Sum$. This motivates considering a fixed vector $\gY$ and the derivative
\begin{align}
\gradq \Sum \gY &= \grad_\qd  \dot{\Sum} \gY \\ 
&=\grad_\qd \left[ -(\gV \times) \Sum + \Sum (\gV \times) \right] \gY
\end{align}
by using \eqref{eq:SumDot}. Rearranging the cross products
\be
\gradq \Sum \gY = \grad_\qd \left[ (\Sum \gY \times) \gV - \Sum (\gY \times) \gV  \right]
\ee
This derivative is now  easier to compute since $\gV = \Sum \gS \qd$ is the only term that depends on $\qd$:
\be
\gradq \Sum \gY =(\Sum \gY \times) \Sum \gS - \Sum (\gY \times) \Sum \gS 
\ee
Using identity \eqref{eq:I1} we have:
\begin{equation}
\gradq \Sum \gY = \Sum ( \Sumt \gY \times) \gS
\label{eq:SumIdentity}
\end{equation}
Following an analogous approach but using identity \eqref{eq:I2},  
\begin{equation}
\gradq \Sum\T \gY = \SumtT \icrf{\Sum\T \gY} \gS
\label{eq:SumT_identity}
\end{equation}
While the global approach of describing the inverse dynamics with a compact system-wide matrix expression follows the approach of \cite{JainRodriguez93}, the approach therein required a strict derivative relationship between $\q$ and $\qd$, which is discharged herein to accommodate broader joint types. These above identities \eqref{eq:SumIdentity} and \eqref{eq:SumT_identity} are the key pieces needed to carry out the remaining derivative calculations. 

\newcommand{\galpha}{\mathsf{\alpha}}

\subsection{Derivatives of motion vectors}
The previous building blocks provide a direct route to obtain the derivatives of the body velocities and accelerations.

{\bf Velocity Derivatives:} We use \eqref{eq:SumIdentity} for velocity derivatives:
\be
\gradq \gV = \gradq \Sum \gv = \Sum  (\Sumt  \gv \times) \gS  + \Sum \gradq (\gS \qd)
\ee
where the first term accounts for the derivative of $\Sum$ and the second the derivative of $\gv = \gS \qd$. We define $\gV_p \triangleq \Sumt \gv$ (stacking the velocity for the predecessor of each body ${}^i \bv_{p(i)}$) and $\gPsid \triangleq \gV_p \times \gS + \gradq (\gS \qd)$. Then, we have that
\begin{equation}
\gradq \gV = \Sum \gPsid \label{eq:vel_deriv_wrt_q}
\end{equation}
Each block element of $\gPsid$ is given by $\Ppsid_j = ({}^j \X_{p(j)} \bv_{p(j)} ) \times \S_j + \grad_{\bq_i} \S_i \bqd_i $.
As a result, this single equation \eqref{eq:vel_deriv_wrt_q} compactly specifies the derivatives of all the body velocities w.r.t. all possible configuration changes as:
\be
\frac{\partial \v_i}{\partial \bq_j} = \begin{cases} \XM{i}{j}\Ppsid_j & j \preceq i \\ \mathbf{0} & o/w \end{cases}
\ee
As already used above, $\grad_\qd \gV = \Sum \gS$.\\[-1.25ex]

{\bf Acceleration Derivatives}
Looking at acceleration
\be
\gradq \gA = \gradq \Sum (\gS \qdd + \gV \times \gv + \gSring \qd)
\ee
Following the same approach used for velocities, we have that
\begin{align}
\gradq \gA &= \Sum \crm{\gA_p} \gS - \Sum \crm{\gv} \gradq \gV \nonumber \\ &~~~~+ \Sum \crm{\gV} \gradq \gS \qd + \Sum \gradq (\gS \qdd + \gSring \qd)  \\
&= \nonumber \Sum \crm{\gA_p}\gS - \Sum \crm{\gv} \Sum \gPsid \\ &~~~~+ \Sum \crm{\gV} \gradq \gS \qd + \Sum \gradq (\gS \qdd + \gSring \qd) 
\end{align}

Using identity \eqref{eq:I1} again we have:
\begin{align}
\gradq \gA &=\Sum \crm{\gA_p}\gS - \crm{\gV} \Sum \gPsid + \Sum \crm{\gV_p} \gPsid \nonumber \\ &~~~~+\Sum \crm{\gV} \gradq \gS \qd + \Sum \gradq (\gS \qdd + \gSring \qd) 
\end{align}
defining $\gPsidd \triangleq \crm{\gA_p} \gS + \crm{\gV_p} \gPsid + \crm{\gV} \gradq \gS \qd + \gradq (\gS \qdd + \gSring \qd)$, we then have that
\be
\gradq \gA =\Sum \gPsidd - \crm{\gV} \Sum \gPsid
\ee

Next, we consider the derivative of acceleration w.r.t. $\qd$.
\begin{align}
\grad_\qd \gA &= \grad_\qd \Sum (\crm{\gV} \gS \qd + \gSring \qd) \\
&= \Sum \crm{\gV} \gS - \Sum \crm{\gv} \Sum \gS + \grad_\qd \gSring \qd
\end{align}
Now we can express: $\grad_\qd \gSring \qd = \gSring + (\grad_\qd \gSring) \qd = \gSring + \gradq \gS \qd$. This slight reorganization is helpful since we already need $\gradq \gS \qd$ to form $\gPsid$ for the velocity derivatives.

Using Identity \eqref{eq:I1} to simplify $ \Sum \crm{\gv} \Sum$, we have that
\begin{align}
\grad_\qd \gA &=  \Sum \left[ \crm{\gV} \gS + \gSring + \crm{\gV_p} \gS + \gradq \gS \qd \right] - \crm{\gV} \Sum \gS
\end{align}
We define $\gUpd \triangleq \gSd  + \gPsid $ to get:
\be
\grad_\qd \gA =  \Sum \gUpd - \crm{\gV} \Sum \gS
\ee

\subsection{Derivatives of Forces}

Using identities \eqref{eq:I2} and \eqref{eq:I3} and the previous outcomes, the results below follow directly. There's one common term that repeatedly appears:
\be
\gB(\gM,\gV) \triangleq \left[\crf{\gV}\gM + \icrf{\gM \gV} - \gM \crm{\gV}\right]
\ee
This definition then provides $\gf = \gM \gA + \frac{1}{2}\gB \gV$.
The body force derivatives take the form
\begin{align}
\grad_\q \gf = \grad_\q   \left[ \gM \gA + \crf{\gV} \gM \gV \right] &= \gM \Sum \gPsidd + \gB \Sum \gPsid \\
\grad_\qd \gf = \grad_\qd \left[ \gM \gA + \crf{\gV} \gM \gV \right] &= \gM \Sum \gUpd + \gB \Sum \gS
\end{align}
While the subtree force derivatives can be derived using identity \eqref{eq:SumT_identity} to provide
\begin{align}
\grad_\q   \gF &= \grad_\q \Sum\T \gf 
            = \SumtT \icrf{\gF} \gS  + \Sum\T \gB \Sum \gPsid + \Sum\T \gM \Sum \gPsidd \\
\grad_\qd \gF &= \Sum\T \grad_\qd \gf 
= \Sum\T \gM \Sum \gUpd + \Sum\T \gB \Sum \gS
\end{align}

\subsection{Derivatives of Torques}
Finally, looking at the inverse dynamics, using $\btau = \gS\T \gF$, the previous results, $\Sum = \Sumt - \eye$, and identity~\eqref{eq:I3}:
\begin{align} 
\grad_\q \btau &= (\gradq \gS)\T\gF + \gS\T \SumtT \icrf{\gF} \gS  \\ &~~~+ \gS\T \Sum\T \gB \Sum \gPsid + \gS\T \Sum\T \gM \Sum \gPsidd \nonumber \\
             &= (\gradq \gS)\T\gF + \gS\T \SumtT (\gF \otimes \gS + \gBC \gPsid + \gMC \gPsidd) \label{eq:fo_q}\\&~~~+ (\gMC \gS)\T \Sum \gPsidd + (\gBC\T \gS )\T \Sum \gPsid \nonumber \\[1.5ex]
\grad_\qd \btau &= \gS\T \Sum\T \gM \Sum \gUpd + \gS\T \Sum\T \gB \Sum \gS \\ 
            &= \gS\T \SumtT ( \gMC \gUpd + \gBC \gS)  + (\gMC \gS )\T \Sum \gUpd  + ( \gBC\T \gS )\T \Sum \gS  \label{eq:fo_qd}       
\end{align}

To return to joint-level quantities for use in our recursive algorithm, we consider the sparsity of $\Sum$ in \eqref{eq:fo_q} and \eqref{eq:fo_qd}. When $j\preceq i$  we have
\begin{align}
\frac{\partial \btau_i}{\partial \bq_j} &= \S_i\T \BC_i \XM{i}{j} \Ppsid_j + \S_i\T \MC_i \XM{i}{j} \Ppsidd_j ~(i\ne j)\\
\frac{\partial \btau_i}{\partial \dot{\bq}_j} &= \S_i\T \BC_i \XM{i}{j}  \S_j + \S_i\T \MC_i \XM{i}{j}  \Uupd_j
\end{align}
while in the special case
\begin{equation}
\frac{\partial \btau_i}{\partial \bq_i} = (\grad_{\bq_i} \S_i)\T \F_i + \S_i\T \BC_i \Ppsid_i + \S_i\T \MC_i \Ppsidd_i
\end{equation}
and when $i \prec j$
\begin{align}
\frac{\partial \btau_i}{\partial \bq_j} &= \S_i\T \XMT{j}{i}  ( \icrf{\F_j} \S_j + \BC_j \Ppsid_j + \MC_j \Ppsidd_j) \\
 \frac{\partial \btau_i}{\partial \dot{\bq}_j} &= \S_i\T \XMT{j}{i}  \BC_j \S_j + \S_i\T \XMT{j}{i}  \MC_j \Uupd_j
\end{align}
Pseudocode to obtain all torque derivatives based on these formulas is given in Algorithm~\ref{alg:tau_FO}.

Note that this derivation ultimately requires three new joint-level computations: (1) for a fixed $\mathbf{y}$, we need $\nabla_{\bq_i} \S_i \, \mathbf{y}$, (2) $\nabla_{\bq_i} \Sring_i \, \bqd_i$, and (3) for a fixed $\F$ we need $\nabla_{\bq_i} \S_i\T \, \F$. Since these are mechanism specific, we rely on AD-generated functions for each constraint-embedding joint model. While additional analytical simplification of these joint-level derivatives is likely possible, our subsequent results show that the resulting compute from AD-generated code represents a small portion of the overall compute cost of the algorithm.

\begin{remark}
The inner while loop  (lines \ref{line:while_start}-\ref{line:while_end}) repeatedly applies transformations, which could be eliminated by working in the global frame \cite{carpentier2019pinocchio,carpentier2018analytical,Singh22}. For constraint embedding models, however, the transform itself also accumulates forces and inertias onto parent bodies, representing an additional step needed to generalize world-frame efficiency gains. Though this reduces compute times by roughly 30\% on average across our benchmarks \cite{GRBA}, we omit it here for clarity; the benchmarked implementation matches the algorithm as presented. 
\end{remark}

\renewcommand{\t}{\mathbf{t}}
\begin{algorithm}[t]
\setstretch{1.15}

\small
\caption{Algorithm For First-Order Derivatives of $\ID$}
\begin{algorithmic}[1]  
\REQUIRE $ \q, \,\qd ,\, \qdd,\, {\rm model}$

\STATE $ \v_0 = 0; \, \a_{{0}} = -\a_g  $

\FOR{$i=1$ to $N$}

\STATE  $ \v_i =  \XM{i}{\p{i}} \v_\p{i} +  \S_i \bqd_i$

\STATE $\a_i =  \XM{i}{\p{i}} \a_\p{i} +  \S_i \ddot{\bq}_i + \v_i \times \S_i \bqd_i + \Sring_i \bqd_i$ \\

\STATE $\Ppsid_i  =  (\XM{i}{\p{i}} \v_\p{i})\times \S_i + \grad_{\bq_i} \S_i \qd_i $\\[.5ex] 
\STATE $\Ppsidd_i  =  (\XM{i}{\p{i}} \a_\p{i})\times \S_i + (\XM{i}{\p{i}} \v_\p{i})\times \Ppsid_i$ 
\STATE $~~~~~~~~~ + \v_i \times \grad_{\bq_i} \S_i \bqd_i + \grad_{\bq_i} (\S_i \bqdd_i + \Sring_i \bqd_i)  $ \\[.5ex]
\STATE $ \Uupd_i  =  \v_i \times \S_i + \Sring_i + \Ppsid_i  $\\[.5ex]

\STATE $\IC_i = \I_i$ ~~;~~$ \F_i =  \I_i  \a_i + ( \v_i \times^*) \I_i \v_i  $ \\[.5ex]
\STATE $ \BC_i = [ ( \v_i \times^*) \I_i -  \I_i  ( \v_i \times ) + (\I_i \v_i )\crff ]  $\\[.5ex]


\ENDFOR


\FOR{$i=N$ to $1$}

\STATE $ \t_1 = \IC_i  \S_i $ ~~; ~~$\t_2 = \BC_i \S_i + \IC_i \Uupd_i $\\[.5ex] 

 \STATE $\t_3 = \BC_i  \Ppsid_i + \IC_i \Ppsidd_i  + \F_i \crff \S_i  $~~;~~ $\t_4 =   \BCT{i} \S_i$\\[.5ex] 

\STATE $j = i$ \\

\STATE $\Jac{\btau}{\q}[i , i] = \t_4\T \Ppsid_i +  \t_1\T \Ppsidd_i + (\grad_{\bq_i} \S_i)\T \F_i$ \,;\, $\Jac{\btau}{\qd}[i , i] = \S_i\T \t_2$ \\[1ex]

\WHILE {$p(j) > 0$}  \label{line:while_start}

    \STATE $\t_1 = \XMT{j}{\p{j}} \t_1$ ; $\t_2 = \XMT{j}{\p{j}} \t_2$ \\[1ex]
    \STATE $\t_3 = \XMT{j}{\p{j}} \t_3$ ; $\t_4 = \XMT{j}{\p{j}} \t_4$ \\[.5ex]
    
    \STATE $j=p(j)$\\[.5ex]
    
    \STATE $\Jac{\btau}{\q}[i, j] =\t_4\T \Ppsid_j +  \t_1\T \Ppsidd_j $~~;~~ $\Jac{\btau}{\q}[j , i] = \S_j\T \t_3$ \label{line:dense1}\\[1ex]
    
    \STATE $\Jac{\btau}{\qd}[i, j] = \t_{4}\T \S_j + \t_{1}\T \Uupd_j $ ~~;~~ $\Jac{\btau}{\qd}[j , i] = \S_j\T \t_2$ \label{line:dense2}\\[1ex]

    \ENDWHILE \label{line:while_end}

    \IF {$\p{i} > 0$}
    
        \STATE $\IC_\p{i} \pluseq  \XMT{i}{\p{i}} \, \IC_i \, \XM{i}{\p{i}}$ \\[1ex]
        \STATE $\BC_\p{i} \pluseq  \XMT{i}{\p{i}} \, \BC_i \, \XM{i}{\p{i}}$ ~~;~~ $\F_\p{i} \pluseq \XMT{i}{\p{i}} \F_i $ 

    \ENDIF

\ENDFOR
\RETURN $\frac{\partial \btau }{\partial{\q}},\frac{\partial \btau }{\partial{\qd}}$

\end{algorithmic}

\label{alg:tau_FO}
\end{algorithm}

\section{Results}
\label{sec:results}

\subsection{Robots}
\begin{figure}[tb]
\centering
\includegraphics[width=.8 \columnwidth]{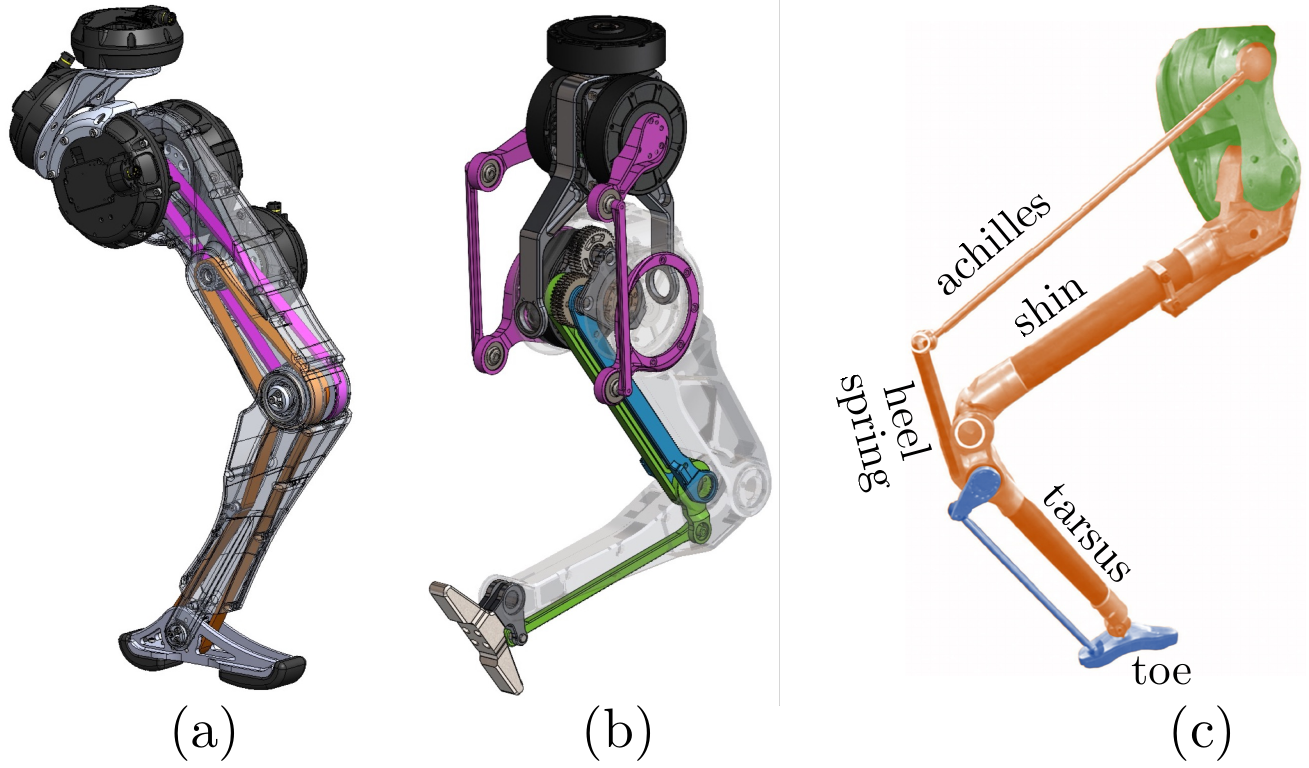}
\caption{Actuation mechanisms for the (a) MIT Humanoid Leg, (b) Tello humanoid  \cite{chignoli2025propagation}, and (c) Cassie Leg. The MIT Humanoid \cite{chignoli2021humanoid} has a coupled belt drive for the knee and ankle. Tello \cite{sim2022tello} has a 2DoF differential hip mechanism and a coupled linkage mechanism for the knee and ankle. Cassie has four bar mechanisms for its distal DoFs.}
\label{fig:systems}
\end{figure}

The accuracy and speed of the algorithm was tested on the KUKA LWR 4+, MIT Mini Cheetah, Agility Robotics' Cassie, MIT Humanoid, Tello biped, and Tello humanoid, with all but the KUKA modeled with constraint embedding. We considered variants of these systems to benchmark the effects of actuation rotors and mechanisms. Both the MIT Mini Cheetah and MIT Humanoid were tested as full models and without motor rotors. For Mini Cheetah, each motor rotor and link formed an aggregate body, yielding a configuration-independent $\S \in \mathbb{R}^{12}$ per joint/rotor pair. For MIT Humanoid, the belts connecting rotors across the hip and knee (Fig.~\ref{fig:systems}(a)) led to a configuration-dependent $\S(\bq) \in \mathbb{R}^{24 \times 2}$ for the knee/ankle aggregate body. 
The Tello biped was tested as an approximate pin-joint model (no mechanisms or rotors), an approximate pin-joint model with rotors, and a full model with motor rotors and actuation transmission mechanisms. This latter case resulted in aggregate bodies for the hip differential mechanisms (pink transmission in Fig.~\ref{fig:systems}(b)) with four rigid bodies each (two rotors, hip roll body, and thigh). The mechanism kinematics account for the transmission of the pink members without modeling them as rigid bodies themselves. The knee-ankle mechanisms (blue and green in Fig.~\ref{fig:systems}(b)) likewise resulted in aggregate bodies with four coupled rigid bodies (two rotors, the shank, and the foot). 
Both mechanisms result in configuration-varying $\S(\bq) \in \mathbb{R}^{24 \times 2}$. For Cassie, two four bar mechanisms drive the ankle and toe joints. The heel spring (orange in Fig.~\ref{fig:systems}(c)) was assumed rigid and attached to the tarsus, such that the orange four bar results in configuration-varying $\S(\bq) \in \mathbb{R}^{18 \times 1}$. In all cases, constraint functions were manually passed to the GRBDA library for it to then generate $\S(\bq)$ and its derivatives via AD.

\subsection{Correctness}
Algorithm~\ref{alg:tau_FO} was tested by extending the GRBDA library \cite{GRBA}.
Its output was validated via comparison with a complex-step approximation \cite{cossette2020complex} of the first-order ID derivatives for the most complete models of the robots previously described. 
The worst-case error (max of: 8.53e-14, Table \ref{tab:robot_sensitivities}) shows near double-precision numerical accuracy. Generally, the velocity derivative $\frac{\partial \btau }{\partial{\qd}}$ is found to have a lower maximum error. 


\begin{table}[!t]
\small
\caption{Maximum Error of the Inverse Dynamics Derivatives}
\centering
\begin{tabular}{ c c c }
\hline
\textbf{Robot} & $\boldsymbol{\partial \tau / \partial q}$ & $\boldsymbol{\partial \tau / \partial \dot{q}}$ \\
\hline
Kuka LWR 4+ & 1.07e-14 & 3.33e-16 \\
Mini Cheetah & 2.84e-14 & 2.66e-15 \\
MIT Humanoid & 2.84e-14 & 1.07e-14 \\
Cassie & 8.53e-14 & 1.11e-14 \\
Tello & 4.26e-14 &  1.78e-15 \\
\hline
\end{tabular}
\label{tab:robot_sensitivities}
\end{table}

\begin{figure}[tb]
\centering
\includegraphics[width=\columnwidth]{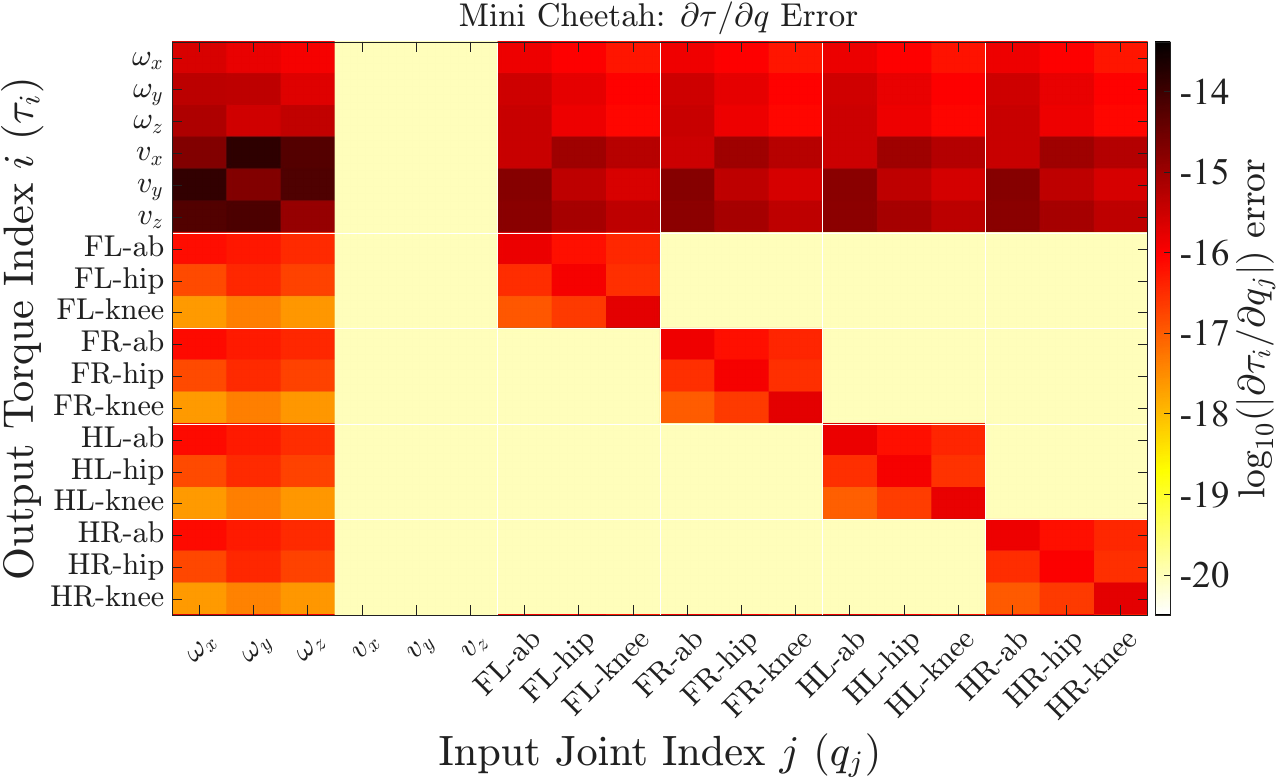}
\caption{Per-joint error for the inverse dynamics derivatives for MIT Mini Cheetah with motor rotors modeled using constraint embedding.}
\label{fig:error_heatmap}
\vspace{-5px}
\end{figure}

The effect of joint location (proximal vs.\ distal) on accuracy was examined by averaging 1000 trials against complex-step on the MIT Mini Cheetah with motor rotors.  Figure \ref{fig:error_heatmap} shows the error in $\frac{\partial \btau_i}{\partial \bq_j}$ for each degree of freedom, where the first 6 correspond to the floating base orientation and position, and the remainder to the leg actuators. The max (over 1000 trials) of the mean entry-wise errors was 1.44e-14, with the largest errors at the floating base wrench. Revolute actuator torque derivatives had lower error, attributed to fewer computational steps; the same trend held for extended serial chains, where errors accumulated for bodies closer to the base due to longer
derivative propagation chains.

\subsection{Runtime}

\begin{figure}[tb]
\centering
\includegraphics[width=1 \columnwidth]{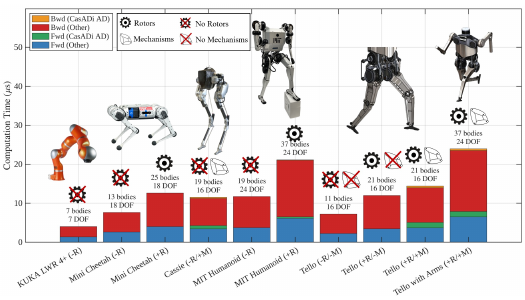}
\caption{Algorithm speed for full and simplified robot models with clang. Fwd refers to the forward pass of Algorithm~\ref{alg:tau_FO} (lines 2-11) and Bwd refers to its backward pass (lines 12-28). $+/-$M indicate with/without actuation mechanism transmission kinematics and $+/-$R indicate with and without motor rotors.}
\label{fig:model_comps}
\end{figure}

The speed of the algorithm was tested on the main robots and their variants, with the average execution time of the forward and backward passes over 1000 trials per robot shown in Fig.~\ref{fig:model_comps}. 
These numbers were obtained following compilation with clang 14 and {\tt -march=native} on a machine running Ubuntu 24 with an Intel i9-14900 CPU (AVX2 compatible) and 62 GiB RAM. 

As expected, the algorithm's computation time increases with more degrees of freedom and more complex joint topologies. However, as shown with the Tello model, the number of rigid bodies modeled had more effect on compute requirements than modeling the mechanism kinematics and treating their derivatives. Specifically, without rotors and mechanisms the approximate Tello model (lower body only) took 7.21 $\mu$s, adding rotors (ten additional rigid bodies) increased compute time to 11.97 $\mu$s (66.0\% increase), while adding the mechanisms only further increased the required time to 14.41 $\mu$s (20.4\% increase). The green and orange bars on Fig.~\ref{fig:model_comps} show the computational cost of derivatives generated by CasADi's AD tools in the case of configuration-dependent $\S(\bq)$. The results show that their current footprint is dominated by the other steps of the algorithm (max of 12.7\% for Tello).

\begin{figure}[tb]
\centering
\includegraphics[width=.95 \columnwidth]{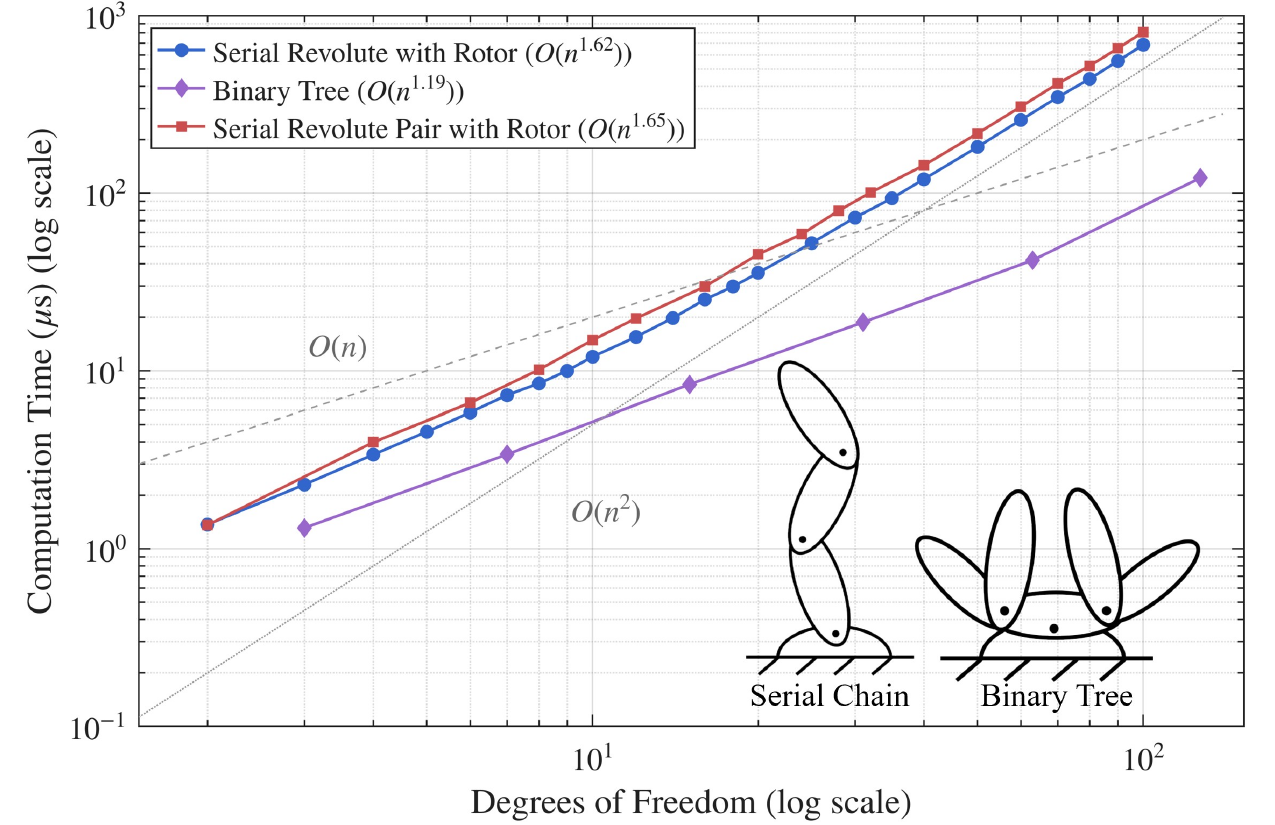}
    \caption{Scaling performance of kinematic trees with different joint types and
    topologies (serial chain vs.~binary tree).}
\label{fig:scaling}
\end{figure}

Two additional tests characterized how chain depth, loop depth, and joint complexity affect
algorithm timing. First, the effect of system size and joint complexity on computation time
was measured for binary-tree and serial-chain systems with
increasing degrees of freedom and different joint types with rotor dynamics, as shown in
Fig.~\ref{fig:scaling}.
The binary tree was tested only with revolute actuators, while the serial chain was tested
with two actuator types: a revolute joint and a pair of revolute joints coupled by a
parallel belt transmission (similar to the MIT Humanoid knee/ankle in
Fig.~\ref{fig:systems}(a)). All cases show computation time scaling between $O(n)$ and
$O(n^2)$; theoretically the algorithm is $O(n^2)$ for uniform chains, but the $O(n)$ terms dominate for the lengths
considered. For a fixed degree-of-freedom count, added actuator coupling increases
computation time. The binary tree requires less computation than the serial chain at equal
degrees of freedom because its reduced depth shortens the inner loop
(lines~\ref{line:while_start}--\ref{line:while_end}).

\begin{figure}[tb]
\centering
\includegraphics[width=.8 \columnwidth]{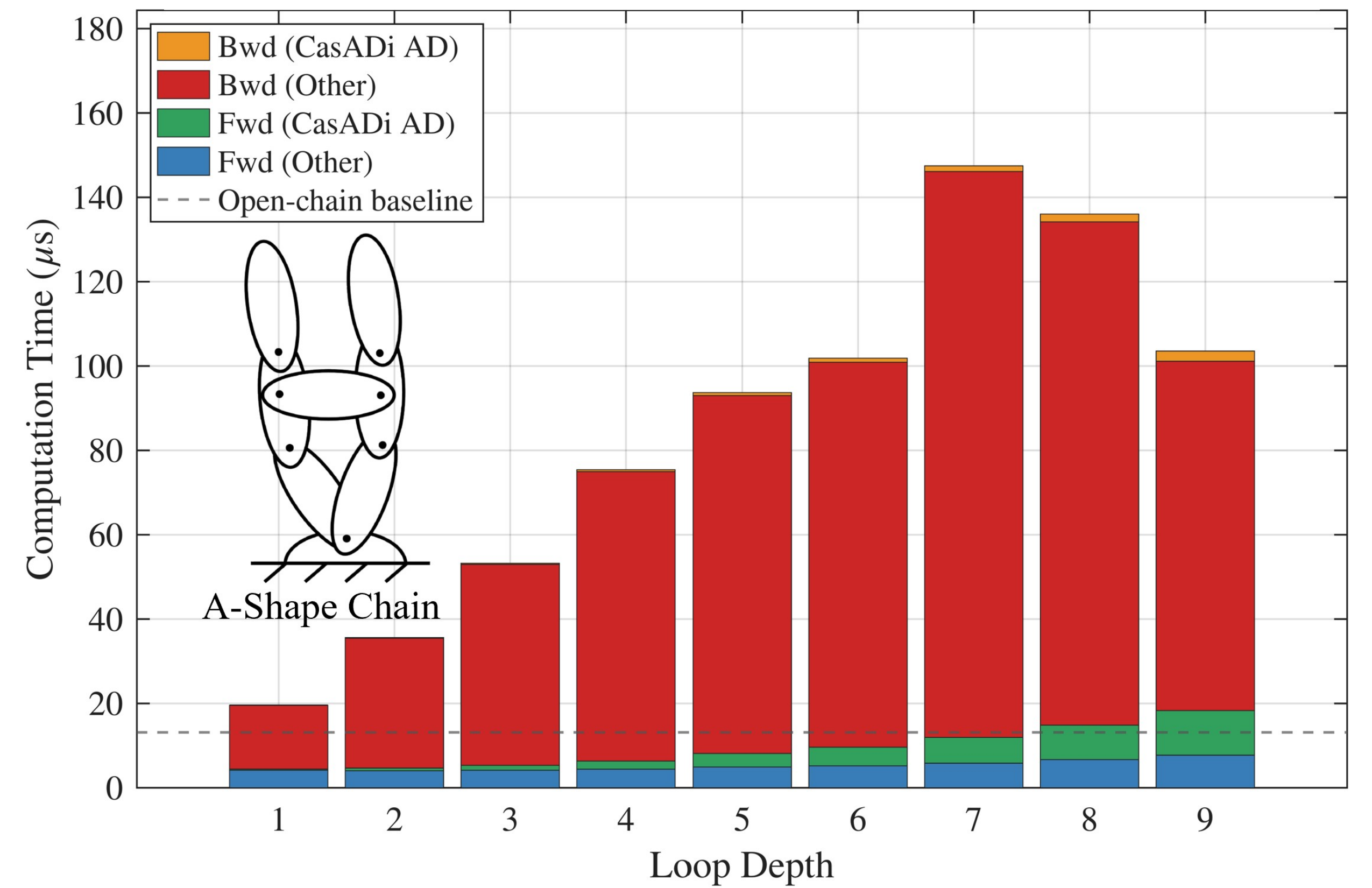}
\caption{Computation time vs.~loop depth for two 10-body chains connected by an additional body at varying depths. Two three link chains connected at a depth of 2 are shown in the inset.}
\label{fig:loop_depth}

\end{figure}

The second test examined the effect of loop size on computation time by coupling two serial
10-body chains (Fig.~\ref{fig:loop_depth}) at increasing depths. Models and constraints were generated automatically via URDF+. Computation time generally increases with loop depth, though it decreases as the loop
approaches the end of the chain. This reduction is attributed to the algorithm's dense
linear algebra on lines~\ref{line:dense1}--\ref{line:dense2}: though matrix sizes grow with loop depth, these lines are executed fewer times, giving batching benefits that can offset the per-operation cost.

\section{Conclusions}
\label{sec:conclusion}

This paper presented a generalization of the derivatives of inverse dynamics to constraint-embedded closed-chain models. By identifying and removing the configuration-invariance assumption on  $\S_i$, we derived adapted first-order derivative formulas that accommodate general joint types, including those arising from constraint embedding. The resulting algorithm was implemented in an open-source library and validated against complex-step approximations across a range of robots.

Our performance analysis showed that actuation kinematics alone introduces minimal overhead relative to conventional pin-joint models, while additional rigid bodies such as motor rotors incur a more substantial cost. We also observed that connectivity topology influences performance, with non-local loops that group many rigid bodies degrading the computational performance of the algorithm.


Several extensions of this work merit future investigation. The derivation could naturally be extended to second-order dynamics derivatives \cite{lee2005newton, singh2024second, nganga2021accelerating}, with GPU acceleration as another possibility \cite{plancher2021accelerating}. 
Second, integrating the generalized derivatives with proximal formulations for contact and non-local constraints~\cite{carpentier2021proximal} could leverage the complementary strengths of constraint embedding and Lagrange multiplier approaches \cite{chignoli2025propagation} when considering forward dynamics derivatives.
For the inverse dynamics derivatives considered herein, exploiting the analytical structure of $\S(\bq)$ could further be useful for treating larger loops. 



\end{document}